\documentclass[runningheads, envcountsame, a4paper]{llncs}
\usepackage{graphicx}
\usepackage{booktabs}       
\usepackage{amsfonts}   
\usepackage{algorithm}
\usepackage[noend]{algpseudocode}

\algrenewcommand\algorithmicforall{\textbf{foreach}}
\algrenewcommand\algorithmicindent{.8em}
\usepackage{caption}
\usepackage{subcaption}
\usepackage{amsmath}
\usepackage{bigints}
\usepackage{xcolor}
\DeclareMathOperator*{\argmax}{argmax}
\usepackage{dsfont}
\usepackage{amsmath}
\usepackage{centernot}
\usepackage[misc]{ifsym}

\begin{document}
\title{MUMBO: MUlti-task Max-value \\Bayesian Optimization\thanks{Supported by EPSRC and the STOR-i Centre for Doctoral Training.}}

\author{Henry B. Moss\inst{1}$^{\textrm{\Letter}}$ \and
David S. Leslie\inst{2}\and
Paul Rayson\inst{3}}

\toctitle{MUMBO: MUlti-task Max-value Bayesian Optimization}
\tocauthor{Henry B. Moss, David S. Leslie, Paul Rayson}

\authorrunning{Moss et al.}
%
\institute{STOR-i Centre for Doctoral Training, Lancaster University \and
Dept. of Mathematics and Statistics, Lancaster University\and
School of Computing and Communications, Lancaster University
\email{h.moss@lancaster.ac.uk}}

\maketitle              
\begin{abstract}

We propose MUMBO, the first high-performing yet computationally efficient acquisition function for multi-task Bayesian optimization. Here, the challenge is to perform efficient optimization by evaluating low-cost functions somehow related to our true target function. This is a broad class of problems including the popular task of multi-fidelity optimization. However, while information-theoretic acquisition functions are known to provide state-of-the-art Bayesian optimization, existing implementations for multi-task scenarios have prohibitive computational requirements. Previous acquisition functions have therefore been suitable only for problems with both low-dimensional parameter spaces and function query costs sufficiently large to overshadow very significant optimization overheads. In this work, we derive a novel multi-task version of entropy search, delivering robust performance with low computational overheads across classic optimization challenges and multi-task hyper-parameter tuning. MUMBO is scalable and efficient, allowing multi-task Bayesian optimization to be deployed in problems with rich parameter and fidelity spaces.

\keywords{Bayesian Optimization  \and Gaussian Processes.}
\end{abstract}

\section{Introduction}
\label{INTRO}
The need to efficiently optimize functions is ubiquitous across machine learning, operational research and computer science. Many such problems have special structures that can be exploited for efficient optimization, for example gradient-based methods on cheap-to-evaluate convex functions, and mathematical programming for heavily constrained problems. However, many optimization problems do not have such clear properties. 

\textbf{Bayesian optimization} (BO) is a general method to efficiently optimize `black-box' functions for which we have weak prior knowledge, typically characterized by expensive and noisy function evaluations, a lack of gradient information, and high levels of non-convexity (see \cite{shahriari2016taking} for a comprehensive review). By sequentially deciding where to make each evaluation as the optimization progresses, BO is able to direct resources into promising areas and so efficiently explore the search space. In particular, a highly effective and intuitive search is achieved through \textbf{information-theoretic} BO, where we seek to sequentially reduce our uncertainty (measured in terms of differential entropy) in the location of the optima with each successive function evaluation \cite{hennig2012entropy,hernandez2014predictive}.

For optimization problems where we can evaluate low-cost functions somehow related to our true objective function,  \textbf{Multi-task} (MT) BO (as first introduced by \cite{swersky2013multi}) provides additional efficiency gains. A popular subclass of MT BO problems is \textbf{multi-fidelity} (MF) BO, where the set of related functions can be meaningfully ordered by their similarity to the objective function. Unfortunately, performing BO over MT spaces has previously required complicated approximation schemes that scale poorly with dimension \cite{swersky2013multi,zhang2017information}, limiting the applicability of information-theoretic arguments to problems with both low-dimensional parameter spaces and function query costs sufficiently large to overshadow very significant optimization overheads. Therefore, MT BO has so far been restricted to considering simple structures at a large computational cost. Despite this restriction, MT optimization has wide-spread use across physical experiments \cite{nguyen2013multidisciplinary,zheng2013multi,pilania2017multi}, environmental modeling \cite{priess2011surrogate}, and operational research \cite{huang2006global,xu2016mo2tos,yong2019multi}. 

For expositional simplicity, this article focuses primarily on examples inspired by tuning the hyper-parameters of machine learning models. Such problems have large environmental impact \cite{strubell-etal-2019-energy}, requiring multiple days of computation to collect even a single (often highly noisy) performance estimate. Consequently, these problems have been proven a popular and empirically successful application of BO \cite{snoek2012practical}. MF applications for hyper-parameter tuning dynamically control the reliability (in terms of bias and noise) of each hyper-parameter evaluation \cite{kennedy2000predicting,lam2015multifidelity,klein2016fast,kandasamy2016gaussian,kandasamy2017multi} and can reduce the computational cost of tuning complicated models by orders of magnitude over standard BO. Orthogonal savings arise from considering hyper-parameter tuning in another MT framework; FASTCV \cite{swersky2013multi} recasts tuning by $K$-fold cross-validation (CV) \cite{kohavi1995study} into the task of simultaneously optimizing the $K$ different evaluations making a single $K$-fold CV estimate. 

Information-theoretic arguments are particularly well suited to such MT problems as they provide a clear measure of the utility (the information gained) of making an evaluation on a particular subtask. This utility then can be balanced with computational cost, providing a single principled decision \cite{swersky2013multi,klein2016fast,mcleod2017practical,zhang2017information}. Despite MT BO being a large sub-field in its own right, there exist only a few alternatives to information-theoretic acquisition functions. Alternative search strategies include extensions of standard BO acquisition functions, including knowledge gradient (KG) \cite{poloczek2017multi,wu2019practical}, expected improvement (EI) \cite{swersky2013multi,picheny2013quantile,lam2015multifidelity}, and upper-confidence bound (UCB) \cite{kandasamy2016gaussian,kandasamy2017multi}. KG achieves  efficient optimization but incurs a high computational overhead. The MT extensions of EI and UCB, although computationally cheap, lack a clear notion of utility and consequently rely on two-stage heuristics, where a hyper-parameter followed by a task are chosen as two separate decisions.  Moreover, unlike our proposed work, the performance of MT variants of UCB and EI depends sensitively on problem-specific parameters which require careful tuning, often leading to poor performance in practical tasks. Information-theoretic arguments have produced the MF BO hyper-parameter tuner FABOLAS \cite{klein2016fast}, out-competing approaches across richer fidelity spaces based on less-principled acquisitions \cite{kandasamy2017multi}. This success motivates our work to provide scalable entropy reduction over MT structures.

We propose \textbf{MUMBO}, a novel, scalable and computationally light implementation of information-theoretic BO for general MT frameworks. Inspired by the work of \cite{wang2017max}, we seek reductions in our uncertainty in the value of the objective function at its optima (a single-dimensional quantity) rather than our uncertainty in the location of the optima itself (a random variable with the same dimension as our search space). MUMBO enjoys three major advantages over current information-theoretic MT approaches:
\begin{itemize}
\item MUMBO has a simple and scalable formulation requiring routine one-dimensional approximate integration, irrespective of the search space dimensions,
\item MUMBO is designed for general MT and MF BO problems across both continuous and discrete fidelity spaces,
\item MUMBO outperforms current information-theoretic MT BO with a significantly reduced computational cost.
\end{itemize}

Parallel work \cite{takeno2019multi} presents essentially the same acquisition function but restricted to discrete multi-fidelity problems from the material sciences. Our article provides a different derivation and general presentation of the method which enables deployment with both discrete and continuous fidelity spaces in general MT BO (including MF). We also provide an implementation in a major BO toolbox and examples across synthetic and hyper-parameter tuning tasks.

\section{Problem Statement and Background}
We now formalize the goal of MT BO, introducing the notation used throughout this work. The goal of  BO is to find the maximzer
\begin{equation}
\textbf{x}^*=\argmax_{\textbf{x}\in\mathcal{X}}g(\textbf{x})
\label{goal}
\end{equation}
of a function $g$ over a $d$-dimensional set of feasible choices $\mathcal{X}\subset\mathds{R}^d$ spending as little computation on function evaluations as possible.

Standard BO seeks to solve (\ref{goal}) by sequentially collecting noisy observations of $g$. By fitting a Gaussian process (GP) \cite{rasmussen2003gaussian}, a non-parametric model providing regression over all functions of a given smoothness (to be controlled by a choice of kernel $k$), we are able to quantify our current belief about which areas of the search space maximize our objective function. An acquisition function $\alpha_n(\textbf{x}):\mathcal{X}\rightarrow \mathds{R}$ uses this belief to predict the utility of making any given evaluation, producing large values at `reasonable' locations. A standard acquisition function \cite{hennig2012entropy} is the expected amount of information provided by each evaluation about the location of the maximum. Therefore after making $n$ evaluations, BO will automatically next evaluate $\textbf{x}_{n+1}=\argmax_{\textbf{x}\in\mathcal{X}}\alpha_n(\textbf{x})$.

\begin{figure*}
     \centering
     \begin{subfigure}[b]{0.31\textwidth}
         \centering
         \includegraphics[width=\textwidth]{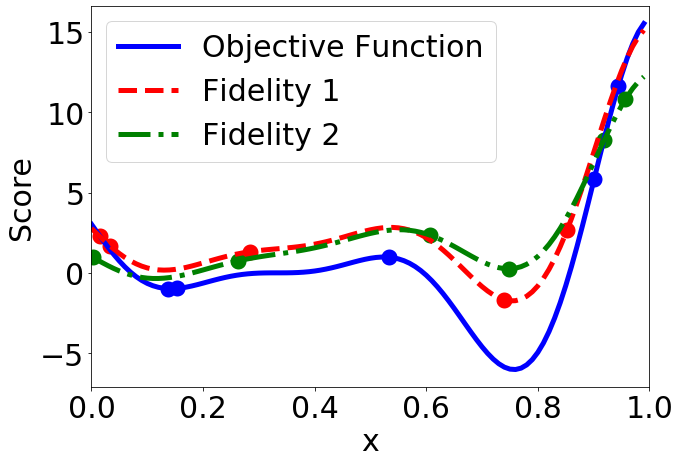}
         \caption{Collected Observations}
     \end{subfigure}
     \begin{subfigure}[b]{0.32\textwidth}
         \centering
         \includegraphics[width=\textwidth]{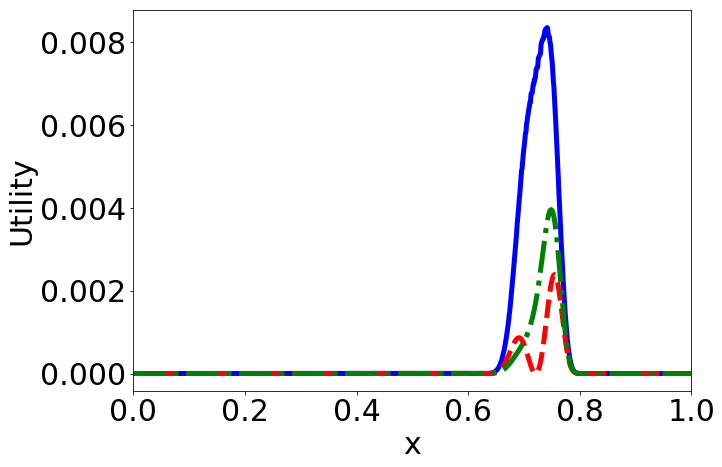}
         \caption{Information gain}
         \label{fig::acq}
     \end{subfigure}
     \begin{subfigure}[b]{0.32\textwidth}
         \centering
         \includegraphics[width=\textwidth]{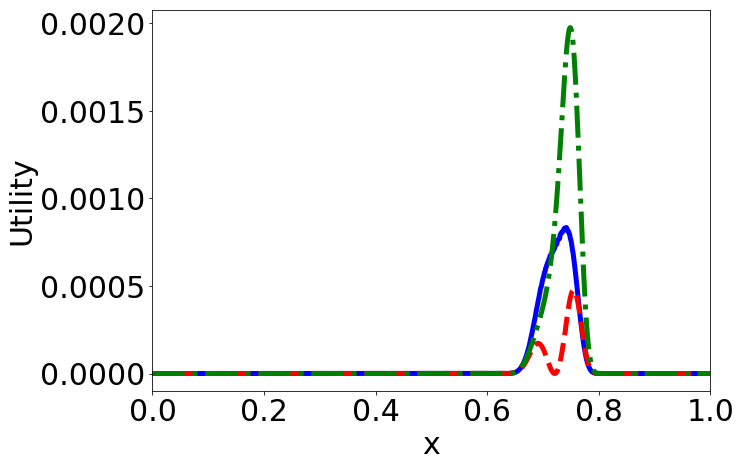}
         \caption{Gain per unit cost}
         \label{fig::costacq}
     \end{subfigure}
        \caption{Seeking the minimum of the 1D Forrester function (blue) with access to two low-fidelity approximations at $\frac{1}{2}$ (red) and $\frac{1}{5}$ (green) the cost of querying the true objective. Although we learn the most from directly querying the objective function, we can learn more per unit cost by querying the roughest fidelity. }
        \label{fig::example}
\end{figure*}

\subsection{Multi-task Bayesian Optimization}
Suppose that instead of querying $g$ directly, we can alternatively query a (possibly infinite) collection of related functions indexed by $\textbf{z}\in\mathcal{Z}$ (henceforth referred to as our fidelity space). We then collect the (noisy) observations $D_n=\{(\textbf{x}_t,\textbf{z}_t,y_t)\}$ for $y_t=f(\textbf{x}_t,\textbf{z}_t)+\epsilon_t$, where $f(\textbf{x},\textbf{z})$ is the result of querying parameter $\textbf{x}$ on fidelity $\textbf{z}$, and $\epsilon_t$ is Gaussian noise. If these alternative functions $f$ are cheaper to evaluate and we can learn their relationship with $g$, then we have access to cheap sources of information that can be used to help find the maximizer of the true task of interest.

\subsection{Multi-task acquisition functions}
The key difference between standard BO and MT BO is that our acquisition function must be able to not only choose the next location, but also which fidelity to evaluate, balancing computational cost with how much we expect to learn about the maximum of $g$. Therefore, we require an extended acquisition function $\alpha_n:\mathcal{X}\times\mathcal{Z}\rightarrow\mathds{R}$ and a cost function $c:\mathcal{X}\times\mathcal{Z}\rightarrow\mathds{R}^+$, measuring the utility and cost of evaluating location $\textbf{x}$ at fidelity $\textbf{z}$ (as demonstrated in Figure \ref{fig::costacq}). In Section \ref{Experiments}, we consider problems both where this cost function is known \textit{a priori} and where it is unknown but estimated using an extra GP \cite{snoek2012practical}. In this work, we seek to make the evaluation that provides the largest information gain per unit cost, i.e. maximizing the ratio
\begin{equation}
    (\textbf{x}_{n+1},\textbf{z}_{n+1})=\argmax_{(\textbf{x},\textbf{z})\in\mathcal{X}\times\mathcal{Z}}\frac{\alpha_n(\textbf{x},\textbf{z})}{c(\textbf{x},\textbf{z})}.
    \label{GOAL}
\end{equation}
\subsection{Multi-task models}
To perform MT BO, our underlying Gaussian process model must be extended across the fidelity space. By defining a kernel over $\mathcal{X}\times\mathcal{Z}$, we can learn predictive distributions after $n$ observations with means $\mu^n(\textbf{x},\textbf{z})$ and co-variances $\Sigma^n((\textbf{x},\textbf{z}),(\textbf{x}',\textbf{z}'))$ from which $\alpha_n(\textbf{x},\textbf{z})$ can be calculated.  Although increasing the dimension of the kernel for $\mathcal{X}$ to incorporate $\mathcal{Z}$ provides a very flexible model, it is argued by \cite{kandasamy2017multi} that overly flexible models can harm optimization speed by requiring too much learning, restricting the sharing of information across the fidelity space. Therefore, it is  common to use more restrictive separable kernels that better model specific aspects of the given problem. 

A common kernel for discrete fidelity spaces is the intrinsic coregionalization kernel of \cite{alvarez2011computationally} (as used in Figure \ref{fig::example}). This kernel defines a co-variance between hyper-parameter and fidelity pairs of
\begin{equation}
    k((\textbf{x},z),(\textbf{x}',z'))=k_{\mathcal{X}}(\textbf{x},\textbf{x}')\times B(z,z'),
    \label{COREG}
\end{equation}
for a base kernel $k_{\mathcal{X}}$ and a positive semi-definite $|\mathcal{Z}|\times|\mathcal{Z}|$ matrix $B$ (set by maximizing the model likelihood alongside the other kernel parameters). $B$ represents the correlation between different fidelities, allowing the sharing of information across the fidelity space. See Section \ref{Experiments} for additional standard MF kernels.

\subsection{Information-theoretic MT BO}
Existing methods for information-theoretic MT BO seek to maximally reduce our uncertainty in the location of the maximizer $\textbf{x}^*=\argmax_{\textbf{x}\in\mathcal{X}}g(\textbf{x})$. Following the work of \cite{hennig2012entropy}, uncertainty in the value of $\textbf{x}^*$ is measured as its differential entropy $H(\textbf{x}^*)=-\mathds{E}_{\textbf{x}\sim p_{\textbf{x}^*}}\left(\log p_{\textbf{x}^*}(\textbf{x})\right)$, where $p_{\textbf{x}^*}$ is the probability density function of $\textbf{x}^*$ according to our current GP model. For MT optimization, we require knowledge of the amount of information provided about the location of $\textbf{x}^*$ from making an evaluation at $\textbf{x}$ on fidelity $\textbf{z}$, measured as the mutual information \begin{equation*}
I(y(\textbf{x},\textbf{z});\textbf{x}^*|D_n)=H(\textbf{x}^*|D_n)-\mathds{E}_{y}\left[H(\textbf{x}^*|y(\textbf{x},\textbf{z}),D_n)\right] 
\end{equation*}
between an evaluation $y(\textbf{x},\textbf{z})=f(\textbf{x},\textbf{z})+\epsilon$ and $\textbf{x}^*$, where the expectation is over $p(y(\textbf{x},\textbf{z})|D_n)$ (see \cite{cover2012elements} for an introduction to information theory). 

Successively evaluating the parameter-fidelity pair that provides the largest information gain per unit of evaluation cost provides the entropy search acquisition function used by \cite{swersky2013multi} and \cite{klein2016fast}, henceforth referred to as the MTBO acquisition function. Unfortunately, the calculation of MTBO relies on sampling-based approximations to the non-analytic distribution of $\textbf{x}^*\,|\,D_n$. Such approximations scale poorly in both cost and performance with the dimensions of our search space (as demonstrated in Section \ref{Experiments}). A modest computational saving can be made for standard BO problems by exploiting the symmetric property of mutual information, producing the predictive entropy search (PES) of \cite{hernandez2014predictive}. However, PES still requires approximations of $\textbf{x}^*\,|\,D_n$ and it is unclear how to extend this approach across MT frameworks.

\section{MUMBO}
\label{MUMBO}
In this work, we extend the computationally efficient information-theoretic acquisition function of \cite{wang2017max} to MT BO. With their max-value entropy-search acquisition function (MES), they demonstrate that seeking to reduce our uncertainty in the value of $g^*=g(\textbf{x}^*)$ provides an equally effective search strategy as directly minimizing the uncertainty in the location $\textbf{x}^*$, but with significantly reduced computation. Similarly, MUMBO seeks to compute the information gain
\begin{align}
\label{MUMBOACQ}
\alpha^{\textit{MUMBO}}_n(\textbf{x},\textbf{z})= &H(y(\textbf{x},\textbf{z})\,|\,D_n)-\mathds{E}_{g^*}\!\!\left[H(y(\textbf{x},\textbf{z})\,|\,g^*\!,D_n)\right],
\end{align} 
which can then be combined with the evaluation cost $c(\textbf{x},\textbf{z})$ (following (\ref{GOAL})). 
Here the expectation is over our current uncertainty in the value of $g^*|D_n$.
\subsection{Calculation of MUMBO}
Although extending MES to MT scenarios retains the intuitive formulation and the subsequent principled decision-making of the original MES, we require a novel non-trivial calculation method to maintain its computational efficiency for MT BO. We now propose a strategy for calculating the MUMBO acquisition function that requires the approximation of only single-dimensional integrals irrespective of the dimensions of our search space.

The calculation of our MUMBO acquisition function (\ref{MUMBOACQ}) for arbitrary $\textbf{x}$ and $\textbf{z}$ must be efficient as each iteration of BO requires a full maximization of (\ref{MUMBOACQ}) over $\textbf{x}$ and $\textbf{z}$ (i.e \ref{GOAL}).  For ease of notation we drop the dependence on $\textbf{x}$ and $\textbf{z}$, so that $g$ denotes the target function value at $\textbf{x}$, $f$ denotes the evaluation of $\textbf{x}$ at fidelity $\textbf{z}$, and $y$ denotes the (noisy) observed value of $f(\textbf{x},\textbf{z})$. Since BO fits a Gaussian process to the underlying functions, our assumptions about $g$ and $y$ imply that their joint predictive distribution is a bivariate Gaussian; with expectation, variance and correlation derived from our GP (as shown in Appendix \ref{appendix::calc}) and denoted by $(\mu_g,\mu_f)$, $(\sigma^2_g,\sigma^2_f+\sigma^2)$ and $\rho$ respectively. These values summarize our current uncertainty in $g$ and $f$ and how useful making an evaluation $y$ will be for learning about $g$. Note that access to this simple two-dimensional predictive distribution is all that is needed to calculate MUMBO (\ref{MUMBOACQ}).

The first term of (\ref{MUMBOACQ}) is the differential entropy of a Gaussian distribution and so can be calculated analytically as $\frac{1}{2}\log(2\pi e(\sigma_f^2+\sigma^2))$. The second term of (\ref{MUMBOACQ}) is an expectation over the maximum value of the true objective $g^*$, which can be approximated using a Monte Carlo approach; we use \cite{wang2017max}'s method to approximately sample a set of $N$ samples $G=\{g_1,\ldots,g_N\}$ from $g^*\,|\, D_n$, using a mean-field approximation and extreme value theory.

It remains to calculate the quantity inside the expectation for a given value of $g^*$. The equivalent quantity in the original MES (without fidelity considerations) was analytically tractable, but we show that for MUMBO this term is intractable. In particular, we show that $y\,|\,g<g^*$ follows an extended-skew Gaussian (ESG) distribution \cite{azzalini1985class,arnold1993nontruncated} in Appendix \ref{appendix::calc}. Unfortunately, \cite{arellano2013shannon} have shown that there is no analytical form for the differential entropy of an ESG. Therefore, after manipulations presented also in Appendix \ref{appendix::calc} and reintroducing dependence on $\textbf{x}$ and $\textbf{z}$, we re-express (\ref{MUMBOACQ}) as 
\vspace{-5pt}
\begin{align}
    \alpha_n^{\textit{MUMBO}}(\textbf{x},\textbf{z})= &\frac{1}{N} \sum\limits_{g^*\in G}\Bigg[\rho(\textbf{x},\textbf{z})^2\frac{\gamma_{g^*}(\textbf{x})\phi(\gamma_{g^*}(\textbf{x}))}{2\Phi(\gamma_{g^*}(\textbf{x}))}-\log(\Phi(\gamma_{g^*}(\textbf{x}))) \nonumber\\&+{\mathds{E}}_{\theta\sim Z_{g^*}(\textbf{x},\textbf{z})}\bigg[\log\Big(\Phi\Big\{\frac{\gamma_{g^*}(\textbf{x})-\rho(\textbf{x},\textbf{z})\theta}{\sqrt{1-\rho^2(\textbf{x},\textbf{z})}}\Big\}\Big)\bigg] \Bigg],
    \label{expanded}
\end{align}
where $\Phi$ and $\phi$ are the standard normal cumulative distribution and probability density functions, $\gamma_{g^*}(\textbf{x})=\frac{g^*-\mu_g(\textbf{x})}{\sigma_g(\textbf{x})}$ and $Z_{g^*}(\textbf{x},\textbf{z})$ is an ESG (with probability density function provided in Appendix \ref{appendix::calc}).

Expression (\ref{expanded}) is analytical except for the final term, which must be approximated for each of the $N$ samples of $g^*$ making up the Monte Carlo estimate. Crucially, this is just a single-dimensional integral of an analytic expression and, hence, can be quickly and accurately approximated using standard numerical integration techniques. We present MUMBO within a BO loop as Algorithm \ref{alg}. 

\begin{algorithm}[ht]
\begin{algorithmic}[1]
\Function{MUMBO}{budget $B$, $N$ samples of $g^*$}
\State Initialize $n\leftarrow 0$, $b\leftarrow 0$
\State Collect initial design $D_0$
\While {$b<B$}
    \State Begin new iteration $n\leftarrow n+1$ 
    \State Fit GP to the collected observations $D_{n-1}$
    \State Simulate $N$ samples of $g^*|D_{n-1}$ 
    \State Prepare $\alpha^{MUMBO}_{n-1}(\textbf{x},\textbf{z})$ as given by Eq. (\ref{expanded})
    \State Find the next point and fidelity to query $(\textbf{x}_n,\textbf{z}_n)\leftarrow\argmax_{(\textbf{x},\textbf{z})}\frac{\alpha^{MUMBO}_{n-1}(\textbf{x},\textbf{z})}{c(\textbf{x},\textbf{z})}$
    \State Collect the new evaluation $y_n\leftarrow f(\textbf{x}_n,\textbf{z}_n)+\epsilon_n$, $\epsilon_n\sim N(0,\sigma^2)$
    \State Append new evaluation to observation set $D_n\leftarrow D_{n-1} \bigcup \{(\textbf{x}_n,\textbf{z}_n),y_n\}$ 
    \State Update spent budget $b\leftarrow b + c(\textbf{x}_n,\textbf{z}_n)$
\EndWhile
\State \Return Believed optimum across $\{\textbf{x}_1,..,\textbf{x}_n\}$
\EndFunction
\end{algorithmic}
\caption{\label{alg}MUlti-fidelity and MUlti-task Max-value Bayesian Optimization: MUMBO}
\end{algorithm}
 
\subsection{Interpretation of MUMBO} 
We provide intuition for (\ref{expanded}) by relating MUMBO to an established BO acquisition function. In the formulation of MUMBO (\ref{expanded}), we see that for a fixed parameter choice $\textbf{x}$ (and ignoring evaluation costs) this acquisition is maximized by choosing the fidelity $z$ that provides the largest $|\rho(\textbf{x},\textbf{z})|$, meaning that the stronger the correlation (either negatively or positively) the more we can learn about the true objective. In fact, if we find a fidelity $\textbf{z}^*$ that provides evaluations that agree completely with $g$, then we would have $\rho(\textbf{x},\textbf{z}^*)=1$ and (\ref{expanded}) would collapse to
\begin{align}
\label{MVES}
    \alpha_n(\textbf{x},\textbf{z}^*)&\nonumber=\frac{1}{N} \sum\limits_{g^*\in G}\bigg[\frac{\gamma_{g^*}(\textbf{x})\phi(\gamma_{g^*}(\textbf{x}))}{2\Phi(\gamma_{g^*}(\textbf{x}))}-\log(\Phi(\gamma_{g^*}(\textbf{x})))\bigg].
\end{align}
This is exactly the same expression presented by \cite{wang2017max} in the original implementation of MES, appropriate for standard BO problems where we can only query the function we wish to optimize.

\subsection{Computational Cost of MUMBO} The computational complexity of any BO routine is hard to measure exactly, due to the acquisition maximization (\ref{GOAL}) required before each function query. However, the main contributor to computational costs are the resources required for each calculation of the acquisition function with respect to problem dimension $d$ and the $N$ samples of $g^*$. Each prediction from our GP model costs $O(d)$, and single-dimensional numerical integration over a fixed grid is $O(1)$. Therefore, a single evaluation of MUMBO can be regarded as an $O(Nd)$ operation. Moreover, as MUMBO relies on the approximation of a single-dimensional integral, we do not require an increase in $N$ to maintain performance as the problem dimension $d$ increases (as demonstrated in Section \ref{Experiments}) and so MUMBO scales linearly with problem dimension. In contrast, the MT BO acquisition used by \cite{swersky2013multi} and \cite{klein2016fast} for information-theoretic MT BO relies on sampling-based approximations of $d$-dimensional distributions, therefore requiring exponentially increasing sample sizes to maintain performance as dimension increases, rendering them unsuitable for even moderately-sized BO problems. In addition, we note that these current approaches require expensive sub-routines and the calculation of derivative information, making their computational cost for even small $d$ much larger than that of MUMBO.

\section{Experiments}
\label{Experiments}

We now demonstrate the performance of MUMBO across a range of MT scenarios, showing that MUMBO provides superior optimization to all existing approaches, with a significantly reduced computational overhead compared to current state-of-the-art. As is common in the optimization literature, we first consider synthetic benchmark functions in a discrete MF setting. Next, we extend the challenging continuous MF hyper-parameter tuning  framework of FABOLAS and use MUMBO to provide a novel information-theoretic implementation of the MT hyper-parameter tuning framework FASTCV, demonstrating that the performance of this simple MT model can be improved using our proposed fully-principled acquisition function. Finally, we use additional synthetic benchmarks to compare MUMBO against a wider range of existing MT BO acquisition functions.

Alongside the theoretical arguments of this paper, we also provide a software contribution to the BO community of a flexible implementation of MUMBO with support for Emukit \cite{emukit2019}. We use a DIRECT optimizer \cite{jones1993lipschitzian} for the acquisition maximization at each BO step and calculate the single-dimensional integral in our acquisition (\ref{expanded}) using Simpson's rule over appropriate ranges (from the known expressions of an ESG's mean and variance derived in Appendix \ref{dervmoments}).

\subsection{General Experimental Details}

Overall, the purpose of our experiments is to demonstrate how MUMBO compares to other acquisition functions when plugged into a set of existing MT problems, focusing on providing a direct comparison with the existing state-of-the-art in information-theoretic MT BO used by \cite{swersky2013multi} and \cite{klein2016fast} (which we name MTBO). Our main experiments also include the performance of popular low-cost MT acquisition functions MF-GP-UCB \cite{kandasamy2016gaussian} and MT expected improvement \cite{swersky2013multi}. In Section \ref{compare} we expand our comparison to include a wider range of existing BO routines, chosen to reflect popularity and code availability. We include the MF knowledge gradient (MISO-KG)\cite{poloczek2017multi}\footnote{As implemented by the original authors at \textit{https://github.com/misokg/NIPS2017}}, an acquisition function with significantly larger computational overheads than MUMBO (and MTBO), as-well as the low-cost acquisition functions of BOCA \cite{kandasamy2017multi} and MF-SKO \cite{huang2006global}. Due to a lack of provided code, and the complexity of their proposed implementations, we were unable to implement multi-fidelity extensions of PES \cite{mcleod2017practical,zhang2017information} or the variant of knowledge-gradient for continuous fidelity spaces \cite{wu2019practical}. As both PES and knowledge gradient require approximations of quantities with dimensionality equal to the search space, their MT extensions will suffer the same scalability issue as MTBO (and MISO-KG). Finally, to demonstrate the benefit of considering MT frameworks,  we also present the standard BO approaches of expected improvement (EI) and max-value entropy search (MES) which query only the true objective. 

To test the robustness of the information-theoretic acquisitions we vary the number of Monte Carlo samples $N$ used for both MUMBO and MTBO (denoted as MUMBO-N and MTBO-N). We report both the time taken to choose the next location to query (referred to as the optimization overhead) and the performance of the believed objective function optimizer (the incumbent) as the optimization progresses. For our synthetic examples, we measure performance after $n$ evaluations as the simple regret $R_n=g(\textbf{x}^*)-g(\hat{\textbf{x}}_n)$, representing the sub-optimality of the current incumbent $\hat{\textbf{x}}_n$. Experiments reporting wall-clock timings were performed on single core Intel Xeon 2.30GHz processors. Detailed implementation details are provided in Appendix \ref{appendix:experiment}.

\subsection{Discrete Multi-fidelity BO}
\label{sec::DiscreteMultiFidelity}

\begin{figure}
     \begin{subfigure}[b]{0.48\textwidth}
         \centering
         \includegraphics[width=\textwidth]{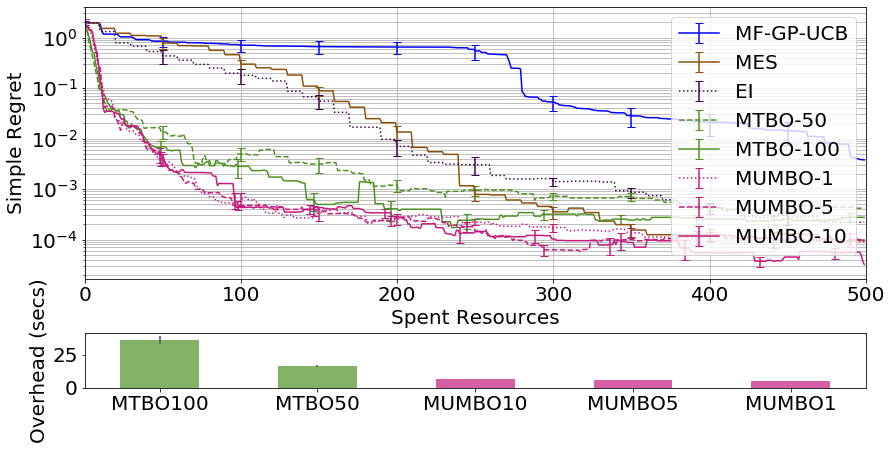}
         \caption{Maximization of the 2D Currin function (2 fidelity levels).}
     \end{subfigure}
     \begin{subfigure}[b]{0.48\textwidth}
         \centering
         \includegraphics[width=\textwidth]{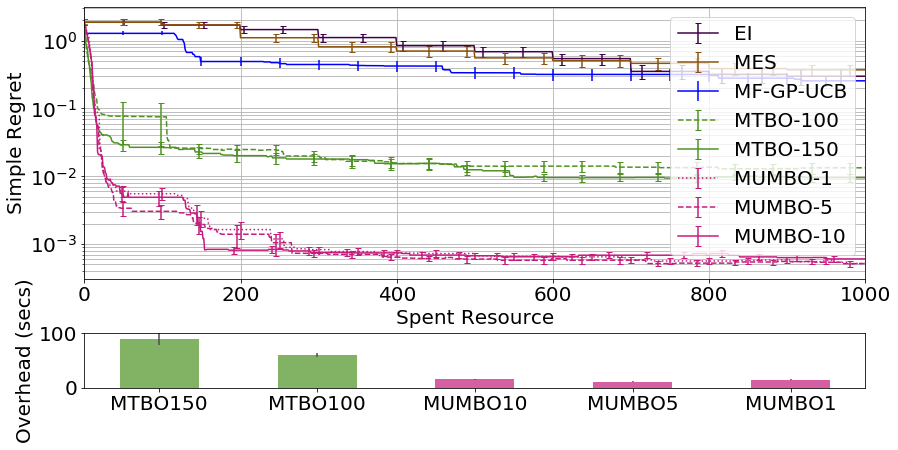}
         \caption{Minimization of 3D Hartmann function (3 fidelity levels).}
     \end{subfigure}
     \begin{subfigure}[b]{0.48\textwidth}
         \centering
         \includegraphics[width=\textwidth]{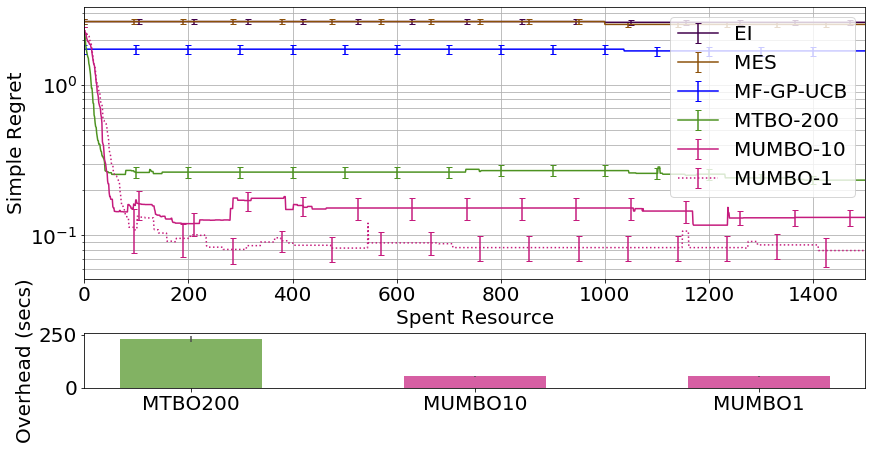}
         \caption{Minimization of 6D Hartmann function (4 fidelity levels).}
     \end{subfigure}
     \begin{subfigure}[b]{0.48\textwidth}
         \centering
         \includegraphics[width=\textwidth]{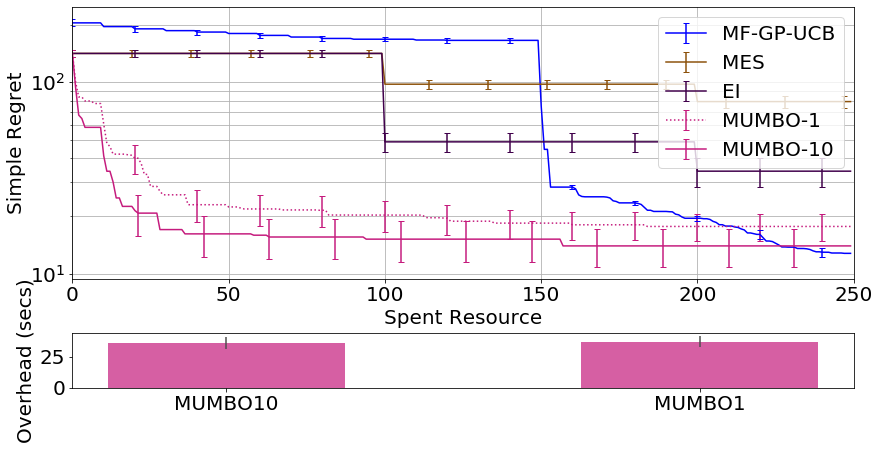}
         \caption{\label{BORE}Maximization of the 8D Borehole function (2 fidelity levels).}
     \end{subfigure}
        \caption{MUMBO provides high-precision optimization with low computational overheads for discrete MF optimization. We show the means and standard errors across $20$ random initializations.}
        \label{discretepics}
\end{figure}

First, we consider the optimization of synthetic problems, using the intrinsic coregionalization kernel introduced earlier (\ref{COREG}). Figure \ref{discretepics} demonstrates the superior performance and light computational overhead of MUMBO across these test functions when we have access to continuous or discrete collections of cheap low-fidelity functions at lower query costs. Although MTBO and MUMBO initially provide comparable levels of optimization, MUMBO quickly provides optimization with substantially higher precision than MTBO and MF-GP-UCB. We delve deeper into the low performance of MF-GP-UCB in Appendix \ref{appendix:discretet}. In addition, MUMBO is able to provide high-precision optimization even when based on a single sample of $g^*$, whereas MTBO requires $50$ samples for reasonable performance on the 2D optimization task, struggles on the 6D task even when based on $200$ samples (requiring $20$ times the overhead cost of MUMBO), and proved computationally infeasible to provide reasonable 8D optimization (and is therefore not included in Figure \ref{BORE}).

Note that MUMBO based on a single sample of $g^*$ is a more aggressive optimizer, as we only consider a single (highly-likely) max-value. Although less robust than MUMBO-10 on average across our examples, this aggressive behavior can allow faster optimization, but only for certain problems (Figure \ref{discretepics}(c)). 
\subsection{Continuous Multi-fidelity BO: FABOLAS}
\label{sec::ContMultiFidelity}
FABOLAS \cite{klein2016fast} is a MF framework for tuning the hyper-parameter of machine learning models whilst dynamically controlling the proportion of available data $z\in(0,1]$  used for each hyper-parameter evaluation. By using the MTBO acquisition and imposing strong assumptions on the structure of the fidelity space, FABOLAS is able to achieve highly efficient hyper-parameter tuning. The use of a `degenerate' kernel \cite{rasmussen2003gaussian} with basis function $\phi(z)=(z,(1-z)^2)^T$ (i.e performing Bayesian linear regression over this basis) enforces monotonicity and strong smoothness across the fidelity space, acknowledging that when using more computational resources, we expect less biased and less noisy estimates of model performance. These assumptions induce a product kernel over the whole space $\mathcal{X}\times\mathcal{Z}$ of:
\begin{equation*}
    k((\textbf{x},z),(\textbf{x}',z'))=k_{\mathcal{X}}(\textbf{x},\textbf{x}')(\phi(z)^T\Sigma_1\phi(z')),
\end{equation*}
where $\Sigma_1$ is a matrix in $\mathds{R}^{2\times2}$ to be estimated alongside the parameters of $k_{\mathcal{X}}$. Similarly, evaluation costs are also modeled in log space, with a GP over the basis $\phi_c(z)=(1,z)^T$ providing polynomial computational complexity of arbitrary degree. We follow the original FABOLAS implementation exactly, using MCMC to marginalize kernel parameters over hyper-priors specifically chosen to speed up and stabilize the optimization. 

\begin{figure}[h]
 \centering
 \captionsetup[subfigure]{width=0.8\textwidth}%
 \begin{subfigure}[t]{0.48\textwidth}
    \includegraphics[width=\textwidth]{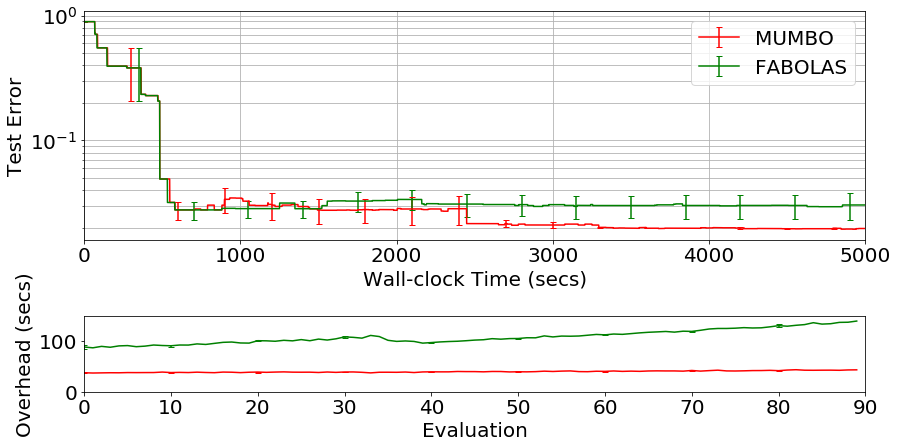}
    \caption{Tuning \textrm{C} and \textrm{gamma} for an SVM to minimize MNIST digit classification error.}
 \end{subfigure}\hfil
 \begin{subfigure}[t]{0.5\textwidth}
    \includegraphics[width=\textwidth]{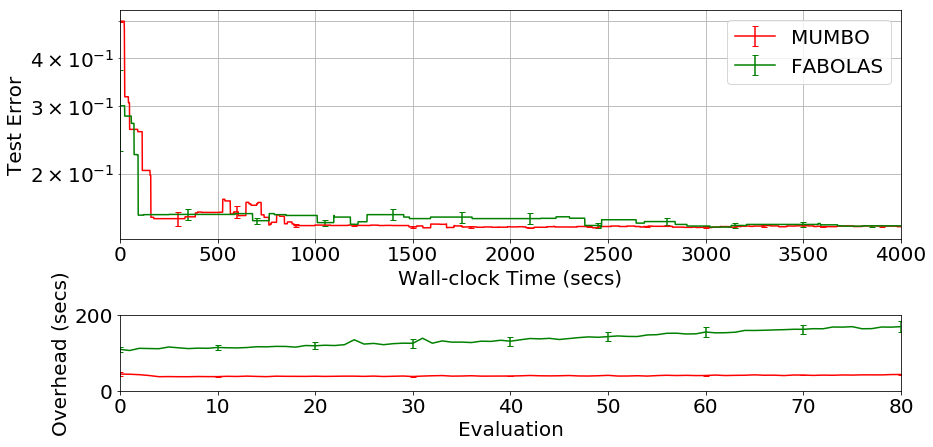}
    \caption{Tuning \textrm{C} and \textrm{gamma} for an SVM to minimize Vehicle Registration classification error.}
 \end{subfigure}
         \caption{MUMBO provides MF hyper-parameter tuning with a much lower overhead than FABOLAS. We show the means and standard errors based on $5$ runs.}
        \label{FABOLASPICS}
\end{figure}

In Figure \ref{FABOLASPICS} we replace the MTBO acquisition used within FABOLAS with a MUMBO acquisition, demonstrating improved optimization on two examples from \cite{klein2016fast}. As the goal of MF hyper-parameter tuning is to find high-performing hyper-parameter configurations after using as few computational resources as possible, including both the fitting of models and calculating the next hyper-parameter and fidelity to query, we present incumbent test error (calculated offline after the full optimization) against wall-clock time. Note that the entire time span considered for our MNIST example is still less than required to try just four hyper-parameter evaluations on the whole data and so we cannot include standard BO approaches in these figures. MUMBO's significantly reduced computational overhead allows twice as many hyper-parameter evaluations as MTBO for the same wall clock time, even though MUMBO consistently queries larger proportions of the data (on average $30\%$ rather than $20\%$ by MTBO). Moreover, unlike MTBO, with an overhead that increases as the optimization progresses, MUMBO remains computationally light-weight throughout and has substantially less variability in the performance of the chosen hyper-parameter configuration. While we do not compare FABOLAS against other hyper-parameter tuning methods, we have demonstrated that, for this well-respected tuner and complicated MF BO problem, that MUMBO provides an improvement in efficiency and a substantial reduction in computational cost.

\subsection{Multi-task BO: FASTCV}
\label{sec::FASTCV}
We now consider the MT framework of FASTCV \cite{swersky2013multi}. Here, we seek the simultaneous optimization of the $K$ performance estimates making up $K$-fold CV. Therefore, our objective function $g$ is the average score across a categorical fidelity space $\mathcal{Z}=\{1,..,K\}$. Each hyper-parameter is evaluated on a single fold, with the corresponding evaluations on the remaining folds inferred using the learned between-fold relationship. Therefore, we can evaluate $K$ times as many distinct hyper-parameter choices as when tuning with full $K$-fold CV whilst retaining the precise performance estimates required for reliable tuning \cite{Moss2018,moss2019fiesta}.

Unlike our other examples, this is not a MF BO problem as our fidelities have the same query cost (at $1/K^{th}$ the cost of the true objective). Recall that all we require to use MUMBO is the predictive joint (bi-variate Gaussian) distribution between an objective function $g(\textbf{x})$ and fidelity evaluations $f(\textbf{x},z)$ for each choice of $\textbf{x}$. For FASTCV, $g$ corresponds with the average score across folds and so (following our earlier notation) our underlying GP provides;
\begin{align}
    \mu_g(\textbf{x})=\frac{1}{K}\sum_{z\in\mathcal{Z}}\mu^n(\textbf{x},z),\nonumber\quad\sigma_g(\textbf{x})^2=\frac{1}{K^2}\sum_{z\in\mathcal{Z}}\sum_{z'\in\mathcal{Z}}\Sigma^n((\textbf{x},z),(\textbf{x},z')),\nonumber
\end{align}
where $\mu^n(\textbf{x},z)$ is the predictive mean performance of $\textbf{x}$ on fold $z$ and $\Sigma^n((\textbf{x},z),(\textbf{x},z'))$ is the predictive co-variance between the evaluations of $\textbf{x}$ on folds $z$ and $z'$ after $n$ hyper-parameter queries. Similarly, we have the correlation between evaluations of $\textbf{x}$ on fold $z$ with the average score $g$ as
\begin{equation*}
    \rho(\textbf{x},z)=\frac{\frac{1}{K}\sum_{z'\in\mathcal{Z}}\Sigma^n((\textbf{x},z),(\textbf{x},z'))}{\sqrt{\sigma^2_g(\textbf{x})\Sigma^n((\textbf{x},z),(\textbf{x},z))}},
\end{equation*}
providing all the quantities required to use MUMBO.
\begin{figure}[ht]
     \centering
      \captionsetup[subfigure]{width=0.8\textwidth}%
     \begin{subfigure}[t]{0.49\textwidth}
         \centering
         \includegraphics[width=\textwidth]{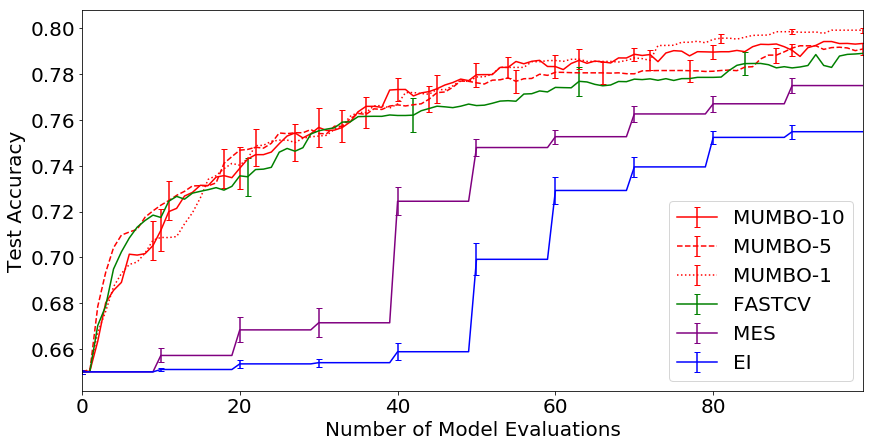}
         \caption{Tuning two SVM hyper-parameters to maximize sentiment classification accuracy for IMDB movie reviews by 10-fold CV.}
     \end{subfigure}
     \begin{subfigure}[t]{0.49\textwidth}
         \centering
         \includegraphics[width=\textwidth]{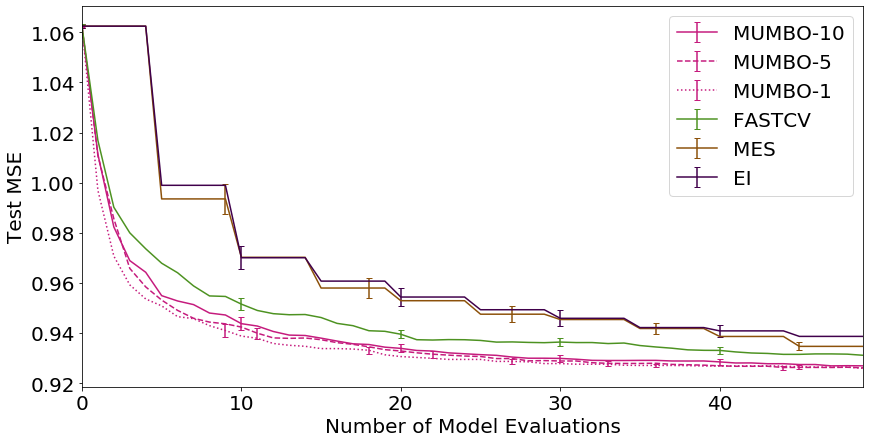}
         \caption{Tuning four hyper-parameters for probabilistic matrix factorization to minimize mean reconstruction error for movie recommendations using 5-fold CV.}
     \end{subfigure}
        \caption{MUMBO provides faster hyper-parameter tuning than the MT framework of FASTCV. We show the mean and standard errors across 40 runs. To measure total computational cost we count each evaluation by $K$-fold CV as $k$ model fits. Experimental details are included in Appendix \ref{Appendix:FASTCV}.}
        \label{FASTCVpics}
\end{figure}

In the original implementation of FASTCV, successive hyper-parameter evaluations are chosen using a two-step heuristic based on expected improvement. Firstly they choose the next hyper-parameter $\textbf{x}$ by maximizing the expected improvement of the predicted average performance and secondly choosing the fold that has the largest fold-specific expected improvement at this chosen hyper-parameter. We instead propose using MUMBO to provide a principled information-theoretic extension to FASTCV. Figure \ref{FASTCVpics} demonstrates that MUMBO provides an efficiency gain over FASTCV, while finding high-performing hyper-parameters substantially faster than standard BO tuning by $K$-fold CV (where we require $K$ model evaluations for each unique hyper-parameter query.

\subsection{Wider Comparison With Existing Methods}
\label{compare}
\begin{figure}
\centering
\captionsetup[figure]{width=0.8\textwidth}%
\begin{minipage}{.48\textwidth}
  \centering
  \includegraphics[width=\textwidth]{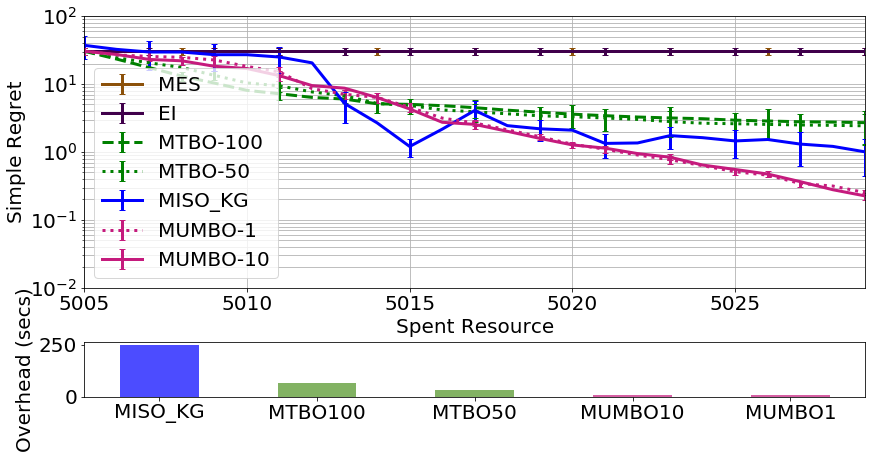}
  \captionof{figure}{The 2D noisy Rosenbrock function (2 fidelities).}
  \label{cts_synth}
\end{minipage}
\begin{minipage}{.48\textwidth}
  \centering
  \includegraphics[width=\textwidth]{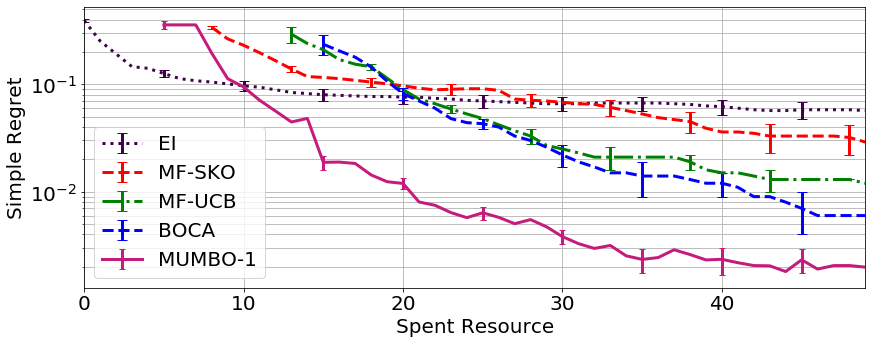}
  \vspace{2pt}
  \captionof{figure}{The 2-d Currin function (1-d continuous fidelity space)}
  \label{UCB_comparison}
\end{minipage}%
\hspace{0\textwidth}

\end{figure}

Finally, we make additional comparisons with existing MT acquisition functions in Figures \ref{cts_synth} and \ref{UCB_comparison}. Knowledge-gradient search strategies are designed to provide particularly efficient optimization for noisy functions, however this high performance comes with significant computational overheads. Although providing reasonable early performance on a synthetic noisy MF optimization task (Figure \ref{cts_synth}), we see that MUMBO is able to provide higher-precision optimization and that, even for this simple 2-d search space, MISO-KG's optimization overheads are magnitudes larger than MUMBO (and MTBO). Figure \ref{UCB_comparison} shows  that MUMBO substantially outperforms existing approaches on a continuous MF benchmark. MF-SKO, MF-UCB and BOCA's search strategies are guided by estimating $g^*$ (rather than $\textbf{x}^*$) and so have comparable computational cost to MUMBO, however, only MUMBO is able to provide high-precision optimization with this low-computational overhead.

\section{Conclusions}
We have derived a novel computationally light information-theoretic approach for general discrete and continuous multi-task Bayesian optimization, along with an open and accessible code base that will enable users to deploy these methods and improve replicability.

MUMBO reduces uncertainty in the optimal value of the objective function with each subsequent query, and provides principled decision-making across general multi-task structures at a cost which scales only linearly with the dimension of the search space. Consequently, MUMBO substantially outperforms current acquisitions across a range of optimization and hyper-parameter tuning tasks.

\newpage
\section{Supplementary Material for MUMBO}
\appendix
\section{Calculation of the MUMBO acquisition function}
\label{appendix::calc}

We now provide a thorough description of our proposed approach to calculate the MUMBO acquisition function for any choice of $\textbf{x}$ and $\textbf{z}$:
\begin{align}
\label{MUMBOACQ2}
\alpha^{MUMBO}_n(\textbf{x},\textbf{z})=H(y(\textbf{x},\textbf{z})\,|\,D_n)-\mathds{E}_{g^*}\!\!\left[H(y(\textbf{x},\textbf{z})\,|\,g^*\!,D_n)\right].
\end{align} 
For ease of notation we drop the dependence on $\textbf{x}$ and $\textbf{z}$, so that $g$ denotes the target function value at $\textbf{x}$, $f$ denotes the evaluation of $\textbf{x}$ at fidelity $\textbf{z}$, and $y$ denotes the (noisy) observed value of $f(\textbf{x},\textbf{z})$, and seek to calculate the respective acquisition value $\alpha_n^{MUMBO}$. From our underlying GP model we can extract our current beliefs about $g$ and $f$ as following a bi-variate Gaussian distribution: 
\begin{eqnarray*}
\begin{pmatrix}g\\
f
\end{pmatrix} & \sim & N\left[\left(\begin{array}{c}
\mu_g\\
\mu_f
\end{array}\right),\left(\begin{array}{ccc}
\sigma_g^2 & \Sigma \\
\Sigma & \sigma_f^2 
\end{array}\right)\right].
\end{eqnarray*}

Then, noting that $Cov(y,g)=\Sigma$, we can write a similar expression for our current beliefs about $g$ and noisy observations $y$ as
\begin{eqnarray*}
\begin{pmatrix}g\\
y
\end{pmatrix} & \sim & N\left[\left(\begin{array}{c}
\mu_g\\
\mu_f
\end{array}\right),\left(\begin{array}{ccc}
\sigma_g^2 & \Sigma \\
\Sigma & \sigma_f^2 + \sigma^2
\end{array}\right)\right].
\end{eqnarray*}

We now derive analytical expressions for these predictive distributions from our underlying GP model. We denote our chosen kernel (defined over $\mathcal{X}\times\mathcal{Z}$) as
$k$, so that $k((\textbf{x},\textbf{z}),(\textbf{x}',\textbf{z}'))$ represents our prior co-variance between the evaluation of $\textbf{x}$ on fidelity $\textbf{z}$ and the evaluation of $\textbf{x}'$ on fidelity $\textbf{z}'$. Denote the location in the fidelity space that corresponds to the true objective function as $\textbf{z}_0$ (i.e. $f(\textbf{x},\textbf{z}_0)=g(\textbf{x})$). For observations $D_n$, let $\textbf{y}_n$ be the observed $y$ values, define the kernel matrix $\textbf{K}_n=\left[k((\textbf{x}_i,\textbf{z}_i),(\textbf{x}_j,\textbf{z}_j))\right]_{(\textbf{x}_i,\textbf{z}_i),(\textbf{x}_j,\textbf{z}_j)\in D_n}$
and kernel vectors $\textbf{k}_n((\textbf{x},\textbf{z}))=\left[k((\textbf{x}_i,\textbf{z}_i),(\textbf{x},\textbf{z}))\right]_{ (\textbf{x}_i,\textbf{z}_i)\in D_n}$. Then, following \cite{rasmussen2003gaussian}, the terms of our bi-variate Gaussian after observations $D_n$ are:
\begin{align*}
    \mu_g=& \textbf{k}_n((\textbf{x},\textbf{z}_0))^T(\textbf{K}_n+\sigma^2I)^{-1}\textbf{y}_n \\
    \mu_f=&  \textbf{k}_n((\textbf{x},\textbf{z}))^T(\textbf{K}_n+\sigma^2I)^{-1}\textbf{y}_n\\
    \sigma^2_g=&k((\textbf{x},\textbf{z}_0),(\textbf{x},\textbf{z}_0))-\textbf{k}_n((\textbf{x},\textbf{z}_0))^T(\textbf{K}_n+\sigma^2I)^{-1}\textbf{k}_n((\textbf{x},\textbf{z}_0))\\
    \sigma^2_f=&k((\textbf{x},\textbf{z}),(\textbf{x},\textbf{z}))-\textbf{k}_n((\textbf{x},\textbf{z}))^T(\textbf{K}_n+\sigma^2I)^{-1}\textbf{k}_n((\textbf{x},\textbf{z}))\\
    \Sigma=&k((\textbf{x},\textbf{z}),(\textbf{x},\textbf{z}_0))-\textbf{k}_n((\textbf{x},\textbf{z}))^T(\textbf{K}_n+\sigma^2I)^{-1}\textbf{k}_n((\textbf{x},\textbf{z}_0))).
\end{align*}
%
Following the advice of \cite{snoek2012practical} we consistently use a Mat\'{e}rn 5/2 kernel to model performance surfaces over the hyper-parameter space.

The first term of (\ref{MUMBOACQ2}) is simply the differential entropy of a Gaussian distribution and so can be calculated analytically as $\frac{1}{2}\log(2\pi e(\sigma_f^2+\sigma^2))$. The second term of (\ref{MUMBOACQ}) is an expectation over the maximum value of the true objective $g^*$, which can be approximated using a Monte Carlo approach; we use \cite{wang2017max}'s method to approximately sample from $g^*\,|\, D_n$ using a mean-field approximation and extreme value theory, generating a set of $N$ values $G=\{g_1,\ldots,g_N\}$. For each $d$-dimensional example in Section \ref{Experiments}, we base our mean-field approximation on a grid of GP predictions at $10,000d$ random locations and any already evaluated locations. Note that we generate only one set of $N$ samples of $g^*$ for each BO step and all the required acquisition queries in that step are calculated with respect to this sample.

All that remains is to calculate the quantity inside the expectation for a given value of $g^*$, i.e the differential entropy of the random variable $y|g<g^*$ with a distribution that we now derive.

\subsection{Derivation of the Extended Skew Normal Distribution}
\label{dervmoments}
To simplify notation, rather than manipulating the co-variance $\Sigma$ directly, we derive MUMBO in terms of the predictive correlation between $y$ and $g$:
\begin{align*}
    \rho=\frac{\Sigma}{\sigma_g\sqrt{\sigma^2_f+\sigma^2}}.
\end{align*}
Then using the well-known result for the conditional distribution of a bi-variate normal, we know that $g\,|\,y$ is also normally distributed with mean $\mu_g+\frac{\sigma_g}{\sqrt{\sigma_f^2+\sigma^2}}\rho(y-\mu_f)$ and variance $\sigma^2_g(1-\rho^2)$. We can therefore write the cumulative distribution function for $y|g\leq g^*$ as

\begin{equation}
\begin{aligned}
\mathds{P}(y\leq \theta | g\leq g*)=&\frac{\mathds{P}(y\leq \theta,g\leq g*)}{\mathds{P}(g\leq g*)}\nonumber\\ 
=&\frac{\bigints_{-\infty}^{\theta}\phi(\frac{u-\mu_f}{\sqrt{\sigma_f^2+\sigma^2}})\Phi\left(\frac{g*-\mu_g-\frac{\sigma_g}{\sqrt{\sigma_f^2+\sigma^2}}
\rho(u-\mu_f)}{\sqrt{\sigma^2_g(1-\rho^2)}}\right)du}{\sqrt{\sigma_f+\sigma^2}\Phi(\frac{g*-\mu_g}{\sigma_g})}.
\end{aligned}
\nonumber
\end{equation}

After differentiating with respect to $\theta$ and defining $\gamma_{g^*}=\frac{g^*-\mu_g}{\sigma_g}$, we can write down the probability density function for the standardized variable $Z_{g^*}=\frac{y-\mu_f}{\sqrt{\sigma_f^2+\sigma^2}}|g<g^*$ as;

\begin{align}
p(\theta)=\frac{1}{\Phi(\gamma_{g^*})}\phi(\theta)\Phi\left(\frac{\gamma_{g^*}-\rho\theta}{\sqrt{1-\rho^2}}\right),
\nonumber
\end{align}
which we recognize as the density of an extended skew normal distribution (ESG) \cite{azzalini1985class}, with moments 
\begin{align}
    \mathds{E}(Z_{g^*})&=\rho\frac{\phi(\gamma_{g^*})}{\Phi(\gamma_{g^*})},\quad{\rm Var}(Z_{g^*})=1-\rho^2\frac{\phi(\gamma_{g^*})}{\Phi(\gamma_{g^*})}\left[\gamma_{g^*}+\frac{\phi(\gamma_{g^*})}{\Phi(\gamma_{g^*})}\right].
    \label{moments}
\end{align}

As \cite{arellano2013shannon} show that the differential entropy of an ESG is non-analytical, so too must be the final term in our MUMBO acquisition (\ref{MUMBOACQ2}). We therefore perform numerical integration using Simpson's rule across eight standard deviations around the mean of $Z_{g^*}$ (quantities provided by (\ref{moments})). Note that the equivalent quantity in the original implementation of MES (without fidelity considerations) has a truncated normal distribution, with a closed form expression for its entropy. 

\subsection{Derivation of the full MUMBO acquisition function}

We can now derive the simplified form (\ref{expanded}) of the MUMBO acquisition function presented in Section \ref{MUMBO}, starting from the information-theoretic definition (\ref{MUMBOACQ2}). Noting that $H(y|g^*,D_n)=H(Z_{g^*})+\frac{1}{2}\log(\sigma_f^2+\sigma^2)$ and that $H(y|D_n)=\frac{1}{2}\log(2\pi e(\sigma_f^2+\sigma^2))$, we can rewrite (\ref{MUMBOACQ2}) for a fixed choice of $\textbf{x}$ and $\textbf{z}$ as
\begin{equation*}
    \alpha_n^{MUMBO}=\frac{1}{2}\log(2\pi e)-\mathds{E}_{g^*}\left[H(Z_{g^*})\right].
\end{equation*}

The differential entropy  $H(Z_{g^*})$ for a fixed sample $g^*$ can be decomposed into three terms
\begin{align}
    \nonumber
    H(Z_{g^*})={\mathds{E}}_{\theta\sim Z_{g^*}}\Bigg[-\log(\phi(\theta))+\log(\Phi(\gamma_{g^*}))-\log\left(\Phi\left(\frac{\gamma_{g^*}-\rho\theta}{\sqrt{1-\rho^2}}\right)\right)\Bigg]
\end{align}

After expanding the first of these terms as
\begin{align}
    {\mathds{E}}_{\theta\sim Z_{g^*}}\left[-\log(\phi(\theta))\right]={\textstyle \frac{1}{2}}{\mathds{E}}_{\theta\sim Z_{g^*}}\left[\theta^2\right]+{\textstyle \frac{1}{2}}\log(2\pi), \nonumber
\end{align}
and further expanding using our expressions for the moments of $Z_{g^*}$, we now have 

\begin{align*}
        \alpha_n^{MUMBO}
    = {\mathds{E}}_{g^*}\Bigg[\rho^2\frac{\gamma_{g^*}\phi(\gamma_{g^*})}{2\Phi(\gamma_{g^*})}-\log(\Phi(\gamma_{g^*}))+ {\mathds{E}}_{\theta\sim Z_{g^*}}\bigg[\log\Big(\Phi\Big\{\frac{\gamma_{g^*}-\rho\theta}{\sqrt{1-\rho^2}}\Big\}\Big)\bigg] \Bigg].
\end{align*}

Therefore, after reintroducing dependence on $\textbf{x}$ and $\textbf{z}$ and replacing the expectation over $g^*$ with a Monte-Carlo approximation across our set of $N$ samples $G$, we see that MUMBO can be expressed as
\begin{align*}
    \alpha_n^{MUMBO}(\textbf{x},\textbf{z})\approx& \frac{1}{N} \sum\limits_{g^*\in G}\Bigg[\rho(\textbf{x},\textbf{z})^2\frac{\gamma_{g^*}(\textbf{x})\phi(\gamma_{g^*}(\textbf{x}))}{2\Phi(\gamma_{g^*}(\textbf{x}))}-\log(\Phi(\gamma_{g^*}(\textbf{x}))) \nonumber\\&+{\mathds{E}}_{\theta\sim Z_{g^*}(\textbf{x},\textbf{z})}\bigg[\log\Big(\Phi\Big\{\frac{\gamma_{g^*}(\textbf{x})-\rho(\textbf{x},\textbf{z})\theta}{\sqrt{1-\rho^2(\textbf{x},\textbf{z})}}\Big\}\Big)\bigg] \Bigg].
\end{align*}

\section{Experiment Details}
\label{appendix:experiment}
We now provide implementation details for our all our experiments. 

\subsection{Discrete Multi-fidelity BO}
\label{appendix:discretet}
Figure \ref{discretepics} shows the performance of MUMBO over the standard MF benchmark functions used by \cite{xiong2013sequential} and \cite{kandasamy2016gaussian}. These problems have an objective function and a discrete hierarchy of low-fidelity approximations that can be  queried at reduced cost. We measure the performance of the MF approaches in terms of the total resources spent on query costs after random initializations. We wish to find high-performing incumbents after spending few resources. We generate starting points for the optimization by querying twice as many random points as the problem dimension and evaluate these across all fidelities. For the information-theoretic approaches we also provide the time spent deciding where to make each successive evaluation as this is an important practical consideration.

In Figure \ref{discretepics} we present the performance of the MF-GP-UCB algorithm of \cite{kandasamy2017multi} using their published code. Unfortunately we were unable to achieve performance on these functions even close to the level claimed in their work. However, our approaches outperform even the results claimed in their paper. This performance discrepancy is likely due to our different initialization scheme and that we do not tune their algorithm's hyper-parameters (illustrating the benefit of using a parameter-free approach like MUMBO). Also note that MF-GP-UCB models fidelities as separate GPs, whereas MUMBO and MTBO use the more sophisticated coregionaliazation model. 

We now provide detailed information about each of our synthetic functions.

\textbf{Forrester Function}. A single dimensional function \cite{forrester2008engineering} defined on $\mathcal{X}=[0,1]$ with three fidelitlies with query costs 10, 5 and 2:
\begin{align*}
    f(x_1,0)&=(6x_1-2)^2\sin(12x_1-4) \\
    f(x_1,1)&= 0.75f(x_1,0)+3(x_1-0.5)+2\\
    f(x_1,2)&= 0.5f(x_1,0)+5(x_1-0.5)+2
\end{align*}

\textbf{Currin exponential function (discete fidelity space)}. A two-dimensional function defined on $\mathcal{X}=[0,1]^2$ with two fidelities queried with costs 10 and 1:
\begin{align*}
    f(x_1,x_2,0)=&\left(1-\exp(-\frac{1}{2x_2})\right)\frac{2300x_1^3+1900x_1^2+2092x_1+60}{100x_1^3+500x_1^2+4x_1+20} \\
    f(x_1,x_2,1)=&\frac{1}{4}f(x_1+0.05,x_2+0.05,0)\\+&\frac{1}{4}f(x_1+0.05,max(0,x_2-0.05),0)  \\+&\frac{1}{4}f(x_1-0.05,x_2+0.05,0)\\+&\frac{1}{4}f(x_1-0.05,max(0,x_2-0.05),0).
\end{align*}

\textbf{Hartmann 3 function}. A three-dimensional function with 4 local extrema defined on $\mathcal{X}=[0,1]^3$ with three fidelities ($m=0,1,2$) queried at costs $100,10$ and $1$:

\begin{align*}
    f(x_1,x_2,&x_3,m) = -\sum\limits_{i=1}^4 \alpha_{i,m+1}\exp\left(-\sum\limits_{j=1}^3A_{i,j}(x_j-P_{i,j})^2\right),
\end{align*}
where 
\begin{align*}
    A&=\begin{pmatrix}
3 & 10 & 30 \\
0.1 & 10 & 35 \\
3 & 10 & 30 \\
0.1 & 10 & 35
\end{pmatrix},
\quad
    \alpha=\begin{pmatrix}
1 & 1.01 & 1.02 \\
1.2 & 1.19 & 1.18  \\
3 & 2.9 & 2.8 \\
3.2 & 3.3 & 3.4
\end{pmatrix},
\quad
    P&=\begin{pmatrix}
3689 & 1170 & 2673 \\
4699 & 4387 & 7470 \\
1091 & 8732 & 5547 \\
381 & 5743 & 8828
\end{pmatrix}.
\end{align*}

\textbf{Hartmann 6 function}. A six-dimensional function defined on $\mathcal{X}=[0,1]^6$ with four fidelities ($m=0,1,2,3$) queried at costs $1000,100,10$ and $1$:

\begin{align*}
    f(x_1,x_2,x_3,x_4,x_5,x_6,m)= -\sum\limits_{i=1}^4\alpha_{i,m+1} \exp\left(-\sum\limits_{j=1}^6A_{i,j}(x_j-P_{i,j})^2\right),
\end{align*}

where 
\begin{align*}
    &A=\begin{pmatrix}
10 & 3 & 17 & 3.5 & 1.7 & 8 \\
0.05 & 10 & 17 & 0.1 & 8 & 14 \\
3 & 3.5 & 1.7 & 10 & 17 & 8\\
17 & 8 & 0.05 & 10 & 0.1 & 14
\end{pmatrix},
\quad
&\alpha=\begin{pmatrix}
1 & 1.01 & 1.02 & 1.03\\
1.2 & 1.19 & 1.18 & 1.17\\
3 & 2.9 & 2.8 & 2.7\\
3.2 & 3.3 & 3.4 & 3.5
\end{pmatrix},
\\
\\
    &P=\begin{pmatrix}
1312 & 1696 & 5569 & 124& 8283& 5886\\
2329 & 4135 & 8307 & 3736& 1004& 9991 \\
2348 & 1451 & 3522 & 2883& 3047& 6650 \\
4047 & 8828 & 8732 & 5743& 1091& 381
\end{pmatrix}.
\end{align*}

\textbf{Borehole function}. An eight-dimensional function defined on \begin{align*}
\mathcal{X}=[&0.05,0.15;100,50,000;63070,115600;990,\\&1110;63.1,116;700,820;1120,1680;9855,12055]
\end{align*} with two fidelities queried with costs 10 and 1:
\begin{align*}
    f(\textbf{x},0)=&\frac{2\pi x_3(x_4-x_6)}{\log(x_2/x_1)\left(1+\frac{2x_7x_3}{log(x_2/x_1)x_1^2x_8}+\frac{x_3}{x_5}\right)}, \\
    f(\textbf{x},1)=&\frac{5x_3(x_4-x_6)}{\log(x_2/x_1)\left(1.5+\frac{2x_7x_3}{log(x_2/x_1)x_1^2x_8}+\frac{x_3}{x_5}\right)}. 
\end{align*}

\subsection{Continuous Multi-fidelity BO: FABOLAS}

For our second set of experiments, we consider the MF hyper-parameter tuning framework of \cite{klein2016fast}, which dynamically chooses the amount of training data used for hyper-parameter evaluations. Their FABOLAS algorithm is widely regarded as state-of-the-art, achieving hyper-parameter tuning with orders of magnitude less computation that standard BO and other competing MF tuning routines. We use the code  provided for FABOLAS within the ROBO package \cite{klein2017robo} by the same authors. We use their implementation exactly, only swapping out their original MTBO acquisition function for our proposed MUMBO acquisition. A good hyper-parameter tuner finds hyper-parameter configurations that will perform well on new data after using as little computational resource as possible, including effort spent fitting models and deciding the hyper-parameter configuration and fidelity to query. By splitting our data into train, validation and test sets, we are able to report  total wall-clock time against the performance (in accuracy) of incumbents on this test set (calculated retrospectively at the end of the optimization). During the optimization, models are trained on random subsets of chosen proportions of the training set and tested on the full validation set.

We consider the same examples as \cite{klein2016fast}, using the same data-sets downloaded from the HPOlib BO benchmark repository \cite{eggensperger2013towards} of MNIST \cite{deng2012mnist} and Vehicle Registrations \cite{siebert1987vehicle} - we refer the reader to their work for specific details. As a result of limited computational resources and wishing to repeat each experiment over multiple random seeds, we had to halve the training data (to $25,000$ for both MNIST and Vehicle) throughout the experiment (including testing the incumbents). We do, however, use the full test and validation sets. For each replication, we start with $10$ random hyper-parameter initializations each evaluated on $\frac{1}{64},\frac{1}{32},\frac{1}{16}$ and $\frac{1}{8}$ of the training data.

\subsection{Multi-task BO: FASTCV}
\label{Appendix:FASTCV}

In Section \ref{sec::FASTCV}, we test MUMBO in a multi-task framework by providing the first information-theoretic implementation of FASTCV \cite{swersky2013multi}, where we sequentially make evaluations on a single $K$-fold CV folds with the aim of optimizing the evaluations based on all $K$ folds. As discussed in Section \ref{sec::FASTCV}, the original implementation of FASTCV chooses hyper-parameters to evaluate and then the fold upon which to make the evaluation as a two-stage heuristic based on the expected improvement acquisition. In Figure \ref{FASTCVpics}, we investigate the change in performance of replacing this acquisition function with the principled MT decision-making provided by our MUMBO acquisition function. We also present the performance of standard BO routines that have to evaluate all $K$ CV folds for each hyper-parameter query.   For ML models, the acquisition function overheads are insignificant compared to the costs of fitting the model on large proportions of the training data (unlike the small proportions chosen by FABOLAS), and so we measure the performance of our algorithms by the number of individual model fits required to reach a certain incumbent performance. To allow the fair comparison of the computational resources used by each algorithm, we consider a single optimization step as the evaluation of a single model on a single fold and so each hyper-parameter evaluation using $K$-fold CV counts as $K$ steps.

We consider two well-known ML tasks: using a support vector machine (SVM) to classify the sentiment in IMDB movie reviews \cite{maas2011learning} and using probabilistic matrix factorization (PMF) \cite{mnih2008probabilistic} to recommend movies on the Movie-lens-100k data set \cite{hoffman2010online}. We tune the regularization strength across $[e^{-5},e^{25}]$ and kernel coefficient across $[e^{-25},e^5]$ for the SVM and the learning rate across $[0,0.01]$, regularization strength across $[0,0.1]$, matrix rank across $[50,..,150]$ and number of model epochs across $[10,..,50]$ for the PMF. To create a difficult MT optimization problem, we use only a small subset of the IMDB data (a random subset of $1,000$ reviews split into $10$ folds) as this increases the between-fold variability of a $K$-fold CV estimate \cite{bengio2004no} and so limits the similarity of evaluations on different folds that is exploited by FASTCV. Despite this challenging MT set-up, both the original FASTCV and MUMBO are able to provide significantly faster tuning than standard approaches, with MUMBO providing an additional increase in test performance over FASTCV (as based on the reliable performance estimates calculated on the $49,000$ reviews not used for training). In addition, we also consider the whole of the large Movelens-100k dataset split into 5 folds. Despite the stochastic nature of PMF meaning that our tuning algorithms have deal with high levels of observation noise for each hyper-parameter evaluation, we once again we see that the principled decision-making of MUMBO allows much faster optimization than all the other approaches - achieving lower $5$-fold CV estimated mean squared error (a standard measurement of performance for recommendation systems).

\subsection{Wider Comparison With Existing Methods}
We now present the functions used for final experiments.

\textbf{Currin exponential function (continuous fidelity space)}. A two-dimensional function defined on $\mathcal{X}=[0,1]^2$ with fidelity space $z\in[0,1]$. The cost of querying fidelity $z$ is given by $\lambda(z)=0.1 +z^2$ with the objective lying at fidelity $z=1$.
\begin{align*}
    f(x_1,x_2,z)=&\left(1-0.1(1-z)\exp(-\frac{1}{2x_2})\right)\frac{2300x_1^3+1900x_1^2+2092x_1+60}{100x_1^3+500x_1^2+4x_1+20}.
\end{align*}

\textbf{Rosenbrock function}. A two-dimensional function defined on $\mathcal{X}=[-2,2]^2$ with two fidelities ($m=0,1$) queried at costs $1000$ and $1$. Observations are contaminated with Gaussian noise with variance $0.001$ and $1e-6$ for each fidelity respectively.

\begin{align*}
    f(x_1,x_2,0)=&(1-x_1)^2+100(x_2-x_1^2)^2\\
    f(x_1,x_2,1)=&f(x_1,x_2,0)+0.1\sin(10x_1+5x_2)
\end{align*}

\end{document}